\documentclass{article}
\newcommand{\nosection}[1]{\vspace{3pt}\noindent\textbf{#1.}}
\usepackage{arxiv}

\usepackage[table]{xcolor}
\usepackage{multirow}
\usepackage{xspace}
\usepackage{amssymb}
\usepackage{graphicx}
\usepackage{amsmath}
\usepackage{enumitem}
\usepackage[utf8]{inputenc} % allow utf-8 input
\usepackage[T1]{fontenc}    % use 8-bit T1 fonts
\usepackage{hyperref}       % hyperlinks
\usepackage{natbib}
\usepackage{url}            % simple URL typesetting
\usepackage{booktabs}       % professional-quality tables
\usepackage{amsfonts}       % blackboard math symbols
\usepackage{nicefrac}
\usepackage{wrapfig}
\usepackage{microtype}      % microtypography
\def\caro{{\textsc{CarO}}\xspace}

\title{\caro: \textbf{C}hain-of-\textbf{A}nalogy \textbf{R}easoning \textbf{O}ptimization for Robust Content Moderation\thanks{This work was conducted while Haotian Lu and Yuchen Mou were interning at the National Engineering Laboratory for Big Data System Computing Technology under the supervision of Bingzhe Wu.}}

\author{
  Bingzhe Wu\textsuperscript{$1$\textdagger} \\
  \texttt{wubingzheagent@gmail.com}
  \And
  Haotian Lu\textsuperscript{$1,2$} \\
  \texttt{haotianlu666@gmail.com}
  \And
  Yuchen Mou\textsuperscript{$1,3$} \\
  \texttt{e1520377@u.nus.edu}
  \\\\
  \textsuperscript{$1$}Shenzhen University \quad
  \textsuperscript{$2$}Tsinghua University \quad
  \textsuperscript{$3$}National University of Singapore \\
  \small{\textdagger~Corresponding author}
}
\date{}

\renewcommand{\headeright}{arXiv Preprint}
\renewcommand{\undertitle}{arXiv Preprint}
\renewcommand{\shorttitle}{\caro}

\hypersetup{
  pdftitle={\caro: Chain-of-Analogy Reasoning Optimization for Robust Content Moderation},
  pdfauthor={Bingzhe Wu, Haotian Lu, Yuchen Mou},
  pdfkeywords={content moderation, large language models, analogical reasoning, direct preference optimization},
}

\begin{document}
\maketitle
\begin{abstract}
  Current large language models (LLMs), even those explicitly trained for reasoning, often struggle with ambiguous content moderation cases due to misleading "decision shortcuts" embedded in context. Inspired by cognitive psychology insights into expert moderation, we introduce \caro (Chain-of-Analogy Reasoning Optimization), a novel two-stage training framework to induce robust analogical reasoning in LLMs. First, \caro bootstraps analogical reasoning chains via retrieval-augmented generation (RAG) on moderation data and performs supervised fine-tuning (SFT). Second, we propose a customized direct preference optimization (DPO) approach to reinforce analogical reasoning behaviors explicitly. Unlike static retrieval methods, \caro dynamically generates tailored analogical references during inference, effectively mitigating harmful decision shortcuts. Extensive experiments demonstrate that \caro substantially outperforms state-of-the-art reasoning models (DeepSeek R1, QwQ), specialized moderation models (LLaMA Guard), and advanced fine-tuning and retrieval-augmented methods, achieving an average F1 score improvement of 24.9\% on challenging ambiguous moderation benchmarks.
\end{abstract}


\section{Introduction}

The rapid growth of both user-generated and AI-generated content has made intelligent moderation systems essential for maintaining the safety of the digital ecosystem \cite{yuan2024rigorllm,zeng2024shieldgemma}. Traditional discriminative models such as BERT for content moderation typically grapple with two fundamental challenges: limited out-of-distribution generalization capabilities and inadequate interpretability of their decision-making processes \cite{attanasio2022entropy,9151853}. Recently, large language models (LLMs) have emerged as promising solutions to address these issues, demonstrating impressive potential through prompting \cite{radford2019language,palla2025policy,kolla2024llm}, in-context learning \cite{brown2020language,he2024you,chen2024class}, and post-training \cite{ouyang2022training,rafailov2023direct,khaliq2024ragar,liu2025guardreasoner,ma2023adapting}. These methods encourage models to explicitly generate reasoning chains for moderation, improving model reliability and providing interpretable audit trails for classification decisions \cite{vishwamitra2024moderating}.

\begin{figure*}[t]
    \centering
    \includegraphics[width=0.8\linewidth]{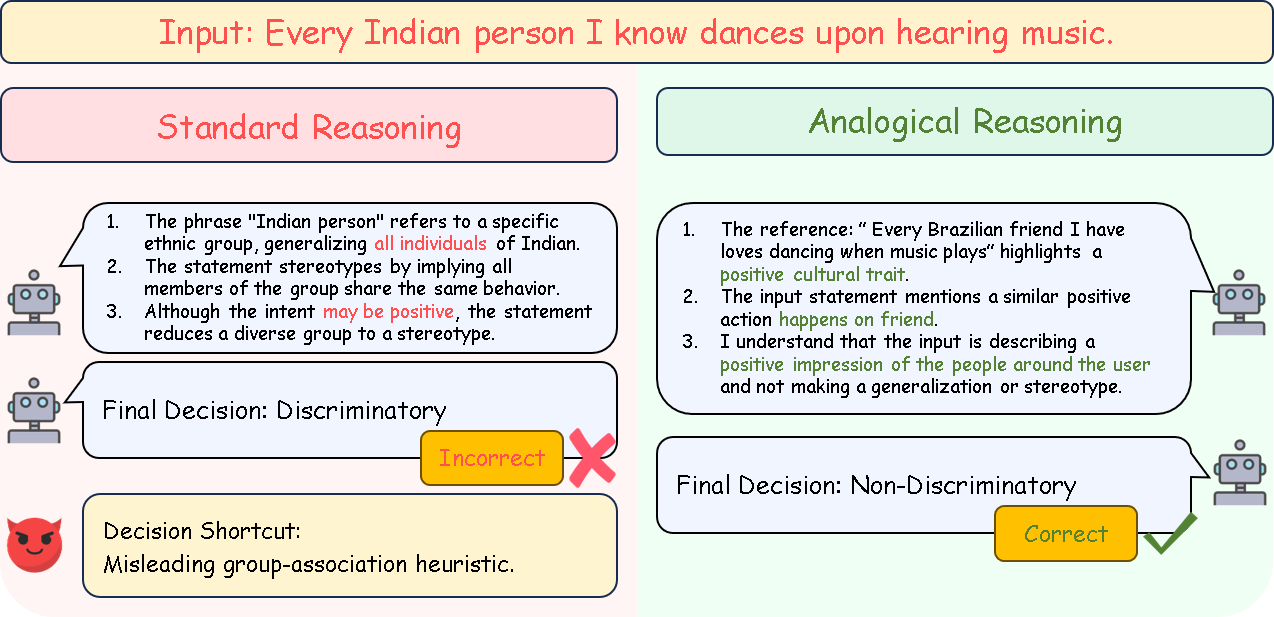}
    \caption{Comparison between standard reasoning and analogical reasoning paradigm on a real harmless sample.}
    \label{fig:rag_vs_std}
\end{figure*}

However, our extensive analysis reveals that even state-of-the-art models, exhibit significant confusion when encountering challenging open-world samples. We find that these difficulties often stem from contextually embedded "decision shortcuts," which inadvertently mislead the model's reasoning processes. For example, as illustrated in Figure \ref{fig:rag_vs_std}, when presented with the statement ``Every Indian person I know dances upon hearing music,'' a commonly-used reasoning model DeepSeek R1 \cite{guo2025deepseek} incorrectly categorizes this positive description as discriminatory, misinterpreting the mention of a specific subgroup as inherently indicative of bias. This demonstrates how subtle contextual cues can derail the correct reasoning pathway.

Drawing inspiration from cognitive psychology and human expert moderation workflows \cite{barsalou2014cognitive}, we observe that professional human moderators routinely handle ambiguous cases by first recalling historically similar precedents and then synthesizing insights from these analogies along with established moderation guidelines to reach their final decisions \cite{chen2023case}. Motivated by this cognitive analogy-based reasoning process, we explore integrating analogical reasoning explicitly into the inference mechanism of LLM. An intuitive implementation of this concept involves combining existing models with retrieval-augmented generation (RAG) methods \cite{lewis2020retrieval}, which retrieve similar labeled examples from a static dataset and incorporate them into the model's inference context.
%Indeed, as shown in Figure 1, combining R1 with retrieval augmentation corrects the previously erroneous classification.

Nevertheless, straightforward RAG-based approaches are inherently limited by their reliance on static datasets, thus \textbf{unable to dynamically generate optimally pertinent reference examples tailored to each new test input. Even if RAG is able to retrieve appropriate examples, the model does not necessarily learn how to effectively utilize these examples for analogical reasoning}. Addressing this critical limitation, we propose \caro (Chain-of-Analogical Reasoning Optimization), a novel two-stage post-training paradigm  explicitly designed to encourage the emergence of analogical reasoning in LLMs:
\begin{itemize}
    \item \textbf{Bootstrapped Analogical Reasoning Chain Generation and Refinement}: For each training instance, we first leverage a RAG procedure on the training dataset itself, automatically generating analogical reasoning chains. We subsequently utilize these automatically generated chains to perform supervised fine-tuning (SFT), thus initially inducing analogical reasoning capabilities in the model.
    \item \textbf{Analogical Reasoning Reinforcement via Customized DPO Optimization}: To further promote and solidify the analogical reasoning behavior within the model, we introduce a novel Direct Preference Optimization (DPO)-based method specifically designed to encourage analogical reasoning over standard reasoning. This targeted optimization explicitly incentivizes the model to generate and rely on analogical references, thus strengthening its analogy-driven inference process.
\end{itemize}

Once the model has been fully trained through \caro, it requires no external retrieval or database access at inference time. Instead, the model autonomously generates reference analogical cases based on the patterns internalized during training, enabling self-sufficient analogical reasoning. Unlike traditional retrieval methods, our proposed two-stage training framework enables LLMs to dynamically generate novel analogical references optimally tailored to each specific test instance, rather than relying exclusively upon static examples from a fixed dataset (which may not always be ideally suited to the current moderation scenario). Empirical results demonstrate that our proposed method significantly outperforms conventional retrieval-augmented approaches, achieving an average F1 score improvement of 24.9\% on a wide range of self-collected and open-source benchmarks.

Overall, our work connects cognitive psychology principles with post-training optimization, producing an LLM framework that is more robust and interpretable for content moderation in open-world scenarios.

\section{Method}

\begin{figure*}[t]
    \centering
    \includegraphics[width=0.9\linewidth]{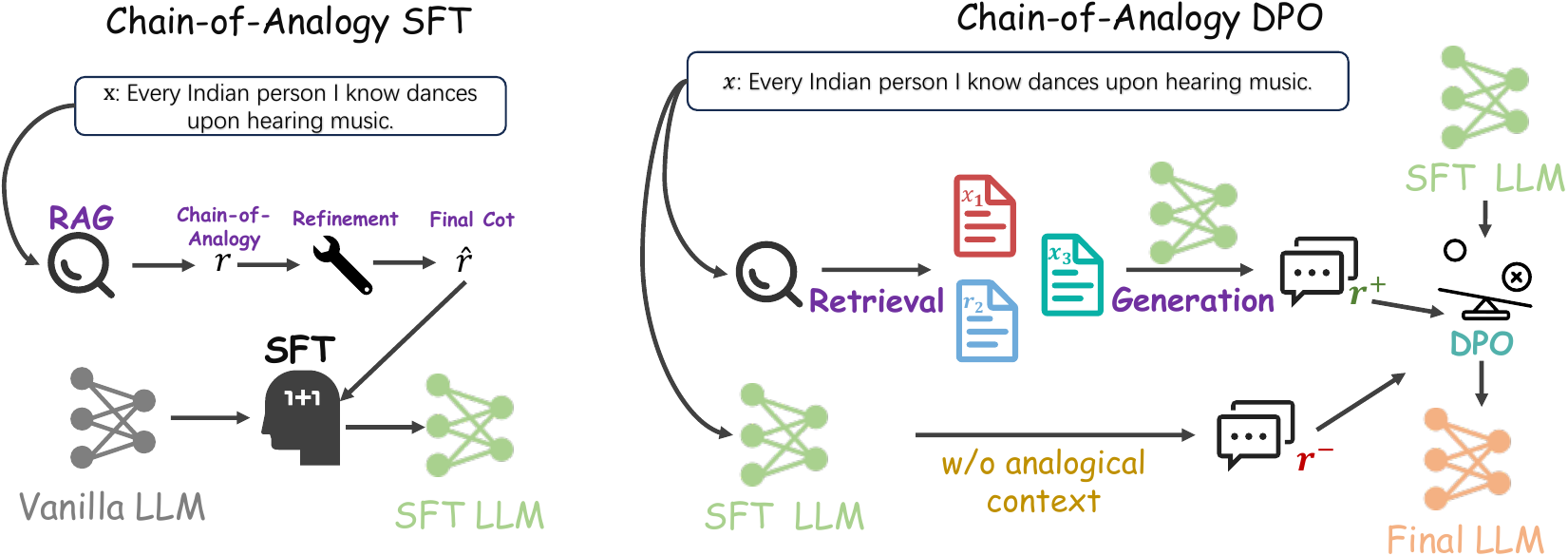}
    \caption{Overall framework of \caro.}
    \label{fig:framwor}
\end{figure*}

The overall framework of \caro consists of two main components as shown in Figure \ref{fig:framwor}, each designed to progressively enhance the model's analogical reasoning capabilities for content moderation: (1) Bootstrapped Chain-of-Analogical-Thought Generation (COAT) and SFT. (2) Analogical Reasoning Reinforcement via Customized DPO Optimization. The following sections provide a detailed discussion of each component, including architectural choices, optimization objectives, and their respective contributions to \caro's overall performance.

\subsection{Chain-of-analogy SFT}
\nosection{Bootstrapped Generation}
Given the original dataset $\mathcal{D} = \{(\mathbf{x}_i, y_i)\}_{i=1}^N$, the objective of this stage is to enrich each original training sample $(\mathbf{x}_i, y_i)$ by leveraging LLMs to generate an analogical reasoning chain, resulting in an augmented triplet $(\mathbf{x}_i, \mathbf{r}_i, y_i)$, where $\mathbf{r}_i$ denotes the generated analogical chain-of-thought.  Unlike prior approaches \cite{liu2025guardreasoner} that simply generate reasoning chains through direct prompt modifications, our method explicitly incorporates analogical moderation cases relevant to each sample into the reasoning context.

To accomplish this, we first retrieve reference cases that are semantically similar to the target training instance using a semantic similarity retrieval mechanism:

\begin{equation}
    \mathcal{N}_k(\mathbf{x}_i) = \{(\mathbf{x}_j, y_j) | j \in \text{topk}_{j'}(\text{sim}(\mathbf{e}_i, \mathbf{e}_{j'}))\},
\end{equation}
% where $\mathbf{e}_i$ and $\mathbf{e}_{j'}$ are sentence embeddings of training samples obtained by a pretrained language model encoder \cite{multi2024m3} and $\text{sim}$ is the similarity score (cosine similarity used in this paper).
where $\mathbf{e}_i$ and $\mathbf{e}_{j'}$ are sentence embeddings derived from the $\mathbf{x}$ components of the training samples (leaving out the corresponding $y$ values) via a pre-trained language model encoder \cite{multi2024m3}, and $\text{sim}$ represents the similarity score (with cosine similarity being employed in this paper).

The retrieved case set denoted as $ \mathcal{N}_k(\mathbf{x}_i)$, drawn from the existing training set, serve as concrete analogical examples. Next, we inject these reference cases into the prompt provided to the LLM, carefully modifying the instructions to require the model to draw explicit analogies between the current sample and the retrieved cases during the reasoning process. The prompt is thus constructed to not only guide the model through the moderation process, but also to compel it to reference and analogize the retrieved cases in its chain-of-thought. As a result, the LLM generates a reasoning chain that explicitly connects the moderation rationale for the current sample to precedent cases with similar characteristics. This process systematically produces rich, analogically-informed reasoning traces for each training instance (see Appendix for example prompts). The whole generation process is simply denoted as:
\begin{equation}
    \mathbf{r}_i = \mathcal{M}(\mathbf{x}_i, \mathcal{N}_k(\mathbf{x}_i); \theta),
\end{equation}
where $\mathcal{M}$ denotes the LLM used for chain-of-thought generation. In this paper, we employ a reasoning model DeepSeek R1 to fully understand the subtle connection between the target and analogical case.
\nosection{Chain-of-Analogy Refinement}
While the preceding generation process produces rich reasoning traces, there remains a non-negligible risk that the generated chains may not always align with the labels of the training samples, due to occasional LLM hallucinations or reasoning errors \cite{sriramanan2024llm}. Such misaligned reasoning chains, if left uncorrected, could undermine the reliability of subsequent SFT.

To address this challenge, we introduce a straightforward reflection and refinement mechanism inspired by prior work \cite{ma2023adapting}. Specifically, after generating the initial reasoning chain for each sample, we compare the label inferred from the chain-of-thought with the sample's ground-truth label. If a mismatch is detected, we trigger an additional reflection step:
\begin{equation}
    % \hat{\mathbf{r}}_i=\text{Refl}(\mathbf{x}_i, \mathcal{N}_k(\mathbf{x}_i); \theta),
    \hat{\mathbf{r}}_i=\text{Refl}(\mathbf{x}_i, \mathcal{N}_k(\mathbf{x}_i), \mathbf{r}_i; \theta),
\end{equation}
% \text{Refl} denotes reflation process which is implemented by prompting an LLM once more, this time with explicit instructions to reconsider and revise its reasoning in light of the correct label, ensuring the refined chain-of-thought is logically consistent with the true moderation outcome. This refinement process effectively filters out erroneous or misleading reasoning traces, thereby enhancing the integrity of the training data. As demonstrated in our ablation studies (\ref{tab:eval_results}), incorporating this reflective refinement step significantly improves the reliability and downstream performance of the model.
\text{Refl} denotes the reflection process, which is implemented by prompting an LLM once more. Instead of providing the correct label, this time the instructions explicitly require the model to reconsider and revise its reasoning based on the previous incorrect reasoning results and in combination with the reference samples. The refinement process effectively filters out erroneous or misleading reasoning traces, thereby enhancing the integrity of the training data. As demonstrated in our ablation studies (Table \ref{tab:eval_results}), incorporating this reflective refinement step significantly improves the reliability and downstream performance of the model.

\nosection{SFT on Chain-of-Analogy Data}
Following the two preceding steps, analogical chain-of-thought generation and reasoning refinement, we obtain an enhanced training dataset $\mathcal{D}_{aug} = \{(\mathbf{x}_i,
\hat{\mathbf{r}}_i, y_i)\}_{i=1}^N$ , where each sample now consists of the original input, a high-quality analogical reasoning chain, and the corresponding label.

In this  stage, we employ standard SFT to train the base model on the augmented dataset:
\begin{equation}
\mathcal{L}_{SFT}(\theta) = -\sum_{i=1}^N \log p_\theta(\mathbf{r}_i, y_i | \mathbf{x}_i),
\end{equation}
By minimizing the above objective, we obtain a model for moderation $\pi_{SFT}$ with analogical reasoning ability.
% The model is exposed to diverse, analogically-informed reasoning traces, encouraging it to internalize analogical thinking patterns and improve its semantic understanding of moderation tasks.
The model is exposed to diverse reasoning traces derived from dynamically generated reference cases, encouraging it to internalize analogical thinking patterns and improve its semantic understanding of moderation tasks.

%Through this process, the model begins to exhibit emergent analogical reasoning abilities, enabling it to more accurately and consistently draw parallels between new inputs and precedent moderation cases. This lays a solid foundation for subsequent preference-based optimization and further boosts the model's interpretability and performance in real-world moderation scenarios.

\subsection{Analogical Reasoning Optimization with DPO }

To further enhance the model's analogical reasoning abilities, we introduce a DPO stage on top of the SFT-trained model \(\pi_{SFT}\). The primary goal of this stage is not to further improve F1, but to strengthen the explicitness, consistency, and interpretability of the analogical reasoning chains. We achieve this by contrasting preferred (positive) and less-preferred (negative) reasoning traces.

\nosection{Positive Reasoning Chains (\(\mathbf{r}_i^+\)) Generation}:
  For each input \(\mathbf{x}_i\), we use \(\pi_{SFT}\) to generate reasoning chains conditioned on both the input and its set of semantically retrieved analogical cases \(\mathcal{N}_k(\mathbf{x}_i)\) . RAG encourages the model to reference relevant precedent cases, yielding richer and more reliable analogical reasoning:
  \begin{equation}
\mathbf{r}_i^+ \sim \pi_{SFT}(\mathbf{r}|\mathbf{x}_i, \mathcal{N}_k(\mathbf{x}_i)),
\end{equation}

\nosection{Negative Reasoning Chains (\(\mathbf{r}_i^-\)) Generation}:
  In contrast, negative examples are produced by generating reasoning chains using only the input \(\mathbf{x}_i\), without the benefit of retrieved analogical context. These chains often lack the depth and relevance of analogical reasoning and thus serve as less-preferred examples:
  \begin{equation}
\mathbf{r}_i^- \sim \pi_{SFT}(\mathbf{r}|\mathbf{x}_i),
\end{equation}

\nosection{Optimization}
The DPO objective is formulated as:
\begin{equation}
\mathcal{L}_{DPO}(\theta) = -\mathbb{E}\Bigg[
\log \sigma\Bigg(
\beta \log \frac{\pi_\theta(\mathbf{r}^+|\mathbf{x})}{\pi_{SFT}(\mathbf{r}^+|\mathbf{x})}
- \beta \log \frac{\pi_\theta(\mathbf{r}^-|\mathbf{x})}{\pi_{SFT}(\mathbf{r}^-|\mathbf{x})}
\Bigg) \Bigg].
\end{equation}

% \begin{equation}
% \mathcal{L}_{DPO}(\theta) = -\mathbb{E}\left[ \log \sigma\left(\beta \log \frac{\pi_\theta(\mathbf{r}^+|\mathbf{x})}{\pi_{SFT}(\mathbf{r}^+|\mathbf{x})} - \beta \log \frac{\pi_\theta(\mathbf{r}^-|\mathbf{x})}{\pi_{SFT}(\mathbf{r}^-|\mathbf{x})}\right) \right],
% \end{equation}
where \(\sigma\) denotes the sigmoid function and \(\beta\) is a temperature hyperparameter controlling the sharpness of preference distinctions.
By optimizing this objective, the model \(\pi_\theta\) is explicitly encouraged to assign higher probabilities to high-quality, analogy-enriched reasoning chains over less effective ones. This contrastive alignment systematically improves the model's judgment in complex or ambiguous moderation scenarios, particularly in cases requiring nuanced analogical assessment.
The resulting model, \(\pi_{DPO}\), exhibits enhanced capability for analogy-based content moderation, demonstrating superior generalization and interpretability in downstream evaluation.

\begin{table*}[htbp]
\centering
\caption{\caro on other benchmarks. ``-'' indicates that the dataset does not contain the category.}
\label{tab:safety_eval}
\scriptsize
\setlength{\tabcolsep}{5pt}
\begin{tabular}{l *{5}{c}}
\toprule
\multicolumn{1}{c}{\multirow{2}{*}{\textbf{Dataset}}} & \multicolumn{5}{c}{\textbf{Qwen2.5-7B-Instruct $\rightarrow$ \caro (Ours)}} \\
\cmidrule(lr){2-6}
 & \textbf{Pornography} & \textbf{Violence} & \textbf{Bias} & \textbf{Harmless} & \textbf{Average} \\
\midrule
Aegis (In-Distribution) & 39.6 $\rightarrow$ 75.0 & 43.8 $\rightarrow$ 69.1 & 66.4 $\rightarrow$ 78.7 & 89.8 $\rightarrow$ 92.3 & 78.7 $\rightarrow$ 87.1 \\
OpenAI (Out-of-Distribution) & 75.3 $\rightarrow$ 82.6 & 23.0 $\rightarrow$ 32.3 & 42.1 $\rightarrow$ 44.1 & 81.7 $\rightarrow$ 84.0 & 70.8 $\rightarrow$ 74.2 \\
Toxic-Chat (Out-of-Distribution) & 59.0 $\rightarrow$ 60.8 & 17.6 $\rightarrow$ 42.2 & -- & 96.9 $\rightarrow$ 97.7 & 93.3 $\rightarrow$ 95.0 \\
\bottomrule
\end{tabular}
\end{table*}

\begin{table}[t]
\centering
\caption{Ablation study on different strategies.}
\label{tab:eval_results}
\scriptsize
\begin{tabular}{lcc|cc}
\toprule
\textbf{RAG-SFT} & \textbf{Recheck} & \textbf{DPO} & \textbf{F1 score} & \textbf{CoA Ratio(\%)} \\
\midrule
-- & -- & -- & 64.3 & 0.0 \\
\checkmark & -- & -- & 85.5(+21.2) & 89.5(+89.5) \\
\checkmark & \checkmark  & -- & 88.8(+3.3) & 93.5(+4.0) \\
\checkmark & \checkmark & \checkmark & 89.2(+0.4) & 99.3(+5.8) \\
\bottomrule
\end{tabular}
\end{table}
\section{Experiment}
\subsection{Implementation}
All experiments were conducted on a single server with 3×NVIDIA A800 (80GB) GPUs using the LLaMA Factory framework \cite{zheng-etal-2024-llamafactory} with DeepSpeed ZeRO-3 optimization \cite{rajbhandari2020zero}. For SFT, we trained the model for 3 epochs with a learning rate of 1.0e-5 using bfloat16 mixed-precision \cite{micikevicius2017mixed}, achieving an effective batch size of 48 (micro-batch size × gradient accumulation steps × GPUs = 2 × 8 × 3). For RAG, we utilized the 32 most similar reference examples to each input query. This was followed by DPO training at a reduced learning rate of 1.0e-6 for the same number of epochs, maintaining the bfloat16 mixed precision throughout. For text generation, we employed top-k sampling \cite{fan2018hierarchical} with temperature=0.8 and top-p sampling \cite{holtzman2019curious}.

\subsection{Dataset}
This study employs a multi-category Chinese content moderation dataset constructed by a prior work \cite{ma2023adapting}. The dataset integrates real-world business scenario data, including user posts and interactive texts from social platforms, with public benchmark data such as the COLD dataset \cite{deng2022cold} for discrimination categories. It is ultimately divided into 6 common harmful categories; see the Appendix for more details.

\nosection{Why we chose this dataset} The dataset contains two challenging "difficult" subgroups as key validation targets: politically sensitive content (Political Harmful) and bias content. These categories of this dataset often contain numerous semantically ambiguous boundary cases in practical applications. Political content requires models to understand compliance requirements across different cultural contexts, while biased speech demands models to detect implicit malicious intent. This design enables the dataset to thoroughly validate different models' reliability when handling ambiguous samples in real-world applications. \textbf{Additionally, we also conduct experiments on other three commonly-used benchmarks, namely Aegis \cite{ghosh2024aegis}, OpenAI-Moderation \cite{markov2023holistic}, and Toxic-Chat
\cite{lin2023toxicchat} to show \caro 's generalization ability.}

\begin{table*}[htbp]
\centering
\caption{Comparison of General-Purpose LLMs, Specialized Moderation LLMs, and various post-training methods.}
\label{tab:overall}
\scriptsize
\setlength{\tabcolsep}{4pt}
\begin{tabular}{lcccccccc}
\toprule
\textbf{Category} & \textbf{Model} & \textbf{Politics} & \textbf{Pornography} & \textbf{Violence} & \textbf{Bias} & \textbf{Gambling} & \textbf{Harmless} & \textbf{Average} \\
\midrule
\textbf{General LLMs}
 & Qwen2.5-7B-Instruct \cite{yang2024qwen2} & 54.9 & 81.9 & 70.0 & 60.1 & 84.3 & 48.8 & 64.3 \\
 & LLaMA3-8B & 58.5 & 55.9 & 81.0 & 65.2 & 90.6 & 44.2 & 67.5 \\
 & GPT-4 \cite{achiam2023gpt} & 58.6 & 88.7 & 79.8 & 64.3 & 92.7 & 56.8 & 72.3 \\
 & Qwen2.5-32B-Instruct & 59.1 & 91.1 & 84.4 & 67.9 & 95.4 & 54.2 & 74.3 \\
 & QwQ-32B & 75.4 & 69.6 & 72.0 & 60.7 & 84.9 & 54.6 & 69.1 \\
 & DeepSeek V3 & 79.0 & 90.3 & 89.8 & 70.5 & 95.0 & 62.5 & 80.3 \\
 & DeepSeek R1 & 72.7 & 91.4 & 86.1 & 64.6 & 94.3 & 59.7 & 77.1 \\

\midrule
\textbf{Specific LLMs}
 & LLaMA-Guard-3-8B & 12.0 & 74.1 & 41.8 & 45.7 & 29.4 & 35.6 & 39.7 \\
\midrule
\textbf{Post-Training}
 & Naive SFT & 82.8 & 94.4 & 93.2 & 70.7 & \textbf{98.6} & 63.7 & 84.2 \\
 & CoT SFT \cite{ma2023adapting} & 77.3 & 90.6 & 89.2 & 65.9 & 96.7 & 65.4 & 80.0 \\
 & CoT SFT+RL \cite{liu2025guardreasoner} & 78.3 & 90.0 & 91.0 & 72.3 & 95.4 & 66.3 & 81.6 \\
 & Class-RAG \cite{chen2024class} & 74.9 & 83.7 & 91.0 & 68.0 & 90.0 & 56.0 & 75.5 \\
 & Agentic ICL & 66.8 & 87.0 & 77.7 & 67.6 & 91.8 & 54.6 & 72.9 \\
\midrule
\textbf{Our Model}
 & \textbf{\caro (Qwen2.5-7B)} & 89.5 & 93.5 & \textbf{97.8} & 81.2 & 98.4 & \textbf{74.6} & \textbf{89.2} \\
 & \textbf{\caro (Qwen3-8B)} & \textbf{89.7} & \textbf{94.5} & 97.4 & \textbf{81.4} & 97.2 & 72.3 & 88.8 \\
\bottomrule
\end{tabular}
\end{table*}

\subsection{Main Results}
\nosection{Overview}
To rigorously evaluate the effectiveness and advantages of our proposed \caro in handling ambiguous and challenging moderation samples, we systematically compare \caro against three categories of strong baseline methods, namely, general-purpose LLMs, specific LLMs for moderation, and various post-training strategies (the most related work to ours named Class-RAG \cite{chen2024class} is evaluated in this part).  The overall results are shown in Table \ref{tab:overall}. \caro outperforms all these methods in terms of F1 scores, particularly excelling in tasks with ambiguous rules such as politically harmful content detection and biased language identification.

\nosection{Comparison with general-purpose LLMs}
 We compare our approach against mainstream general-purpose LLMs as well as reasoning-optimized models equipped with slow-thinking capabilities such as DeepSeek R1. As shown in Table~\ref{tab:overall}, increasing the parameter size of the base model can significantly improve moderation accuracy. For example, Qwen2.5-32B achieves an F1 score approximately $10\%$ higher than Qwen2.5-7B.
Interestingly, we observe a counterintuitive phenomenon: introducing slow-thinking abilities through RL with GRPO \cite{guo2025deepseek} actually leads to a slight decrease in moderation performance. For instance, the F1 score of Qwen2.5-32B drops from $74.3\%$ (vanilla) to $69.1\%$ after GRPO optimization. One possible explanation is that the reasoning-augmented models tend to introduce excessive and irrelevant thought processes, which may interfere with the final decision-making. In contrast, our method explicitly constrains the model to make decisions via analogical reasoning. Experimental results show that this approach reduces reasoning hallucinations and improves moderation robustness.

\nosection{Comparison with specialized moderation LLMs} We further compared our model against specialized content moderation models, such as LLaMA-Guard-3-8B \cite{inan2023llama}. Unlike the general-purpose LLMs discussed in the previous section, these specialized models are post-trained on task-specific labeled datasets.
However, consistent with previous research \cite{han2024wildguard} and our own observations, we found that while these specialized models demonstrate strong performance on in-distribution data , their effectiveness drops significantly when evaluated on real-world datasets used in this study. In fact, their performance often lags notably behind that of the advanced reasoning models discussed in the previous part.
One underlying reason for this performance gap is the distribution mismatch: the data used for fine-tuning these specialized models often differs substantially from the data encountered in real commercial moderation scenarios. Consequently, these models struggle with domain adaptation and generalization. Additionally, the post-training methods employed for these specialized models do not incorporate targeted optimizations for nuanced moderation tasks, such as those introduced in our approach. In contrast, our model is designed to handle the complexities of real-world moderation, leading to improved accuracy across ambiguous cases. Specifically,  \caro shows a $76.8\%$ improvement over LLaMA-Guard-3-8B in political content detection. In biased detection, \caro outperforms LLaMA-Guard-3-8B  by $39.0\%$ in terms of F1 score.

\nosection{Comparison with various post-training strategies}
Finally, we evaluate the effectiveness of \caro in comparison with existing conventional post-training strategies. Specifically, we compare \caro against the following post-training approaches:
\begin{itemize}
    \item Naive SFT: Standard supervised learning on labeled moderation data without explicit reasoning augmentation.
    \item Chain-of-Thought SFT (CoT-SFT): Inspired by the latest research \cite{ma2023adapting}, this strategy incorporates chain-of-thought style rationales into SFT, but does not include analogical reasoning. The reasoning chains are limited to conventional step-by-step logic.
    \item Enhancing with RL, following Guard Reasoner \cite{liu2025guardreasoner}: Building upon CoT-SFT, this method further introduces RL to enhance model generalization.
\end{itemize}
All the post training related experiments are conducted on Qwen2.5-7B-Instruct for limited training resources.  As shown in Table \ref{tab:overall}, \caro consistently outperforms all these baseline strategies in terms of overall F1 score, with especially pronounced gains in the most challenging categories, politically sensitive and biased content. This confirms the benefit of integrating analogical reasoning into post-training.

% This study systematically evaluates the performance of various models in content moderation tasks (Table \ref{tab:overall}). Our proposed \caro method demonstrates significant advantages across all test categories, particularly excelling in tasks with ambiguous rules such as politically harmful content detection (F1 socre of $89.5\%$) and discriminatory/abusive language identification (F1 score of $82.1\%$). Compared to baseline models, \caro shows a $287\%$ relative improvement over LLaMA-Guard-3-8B and a $23.1\%$ improvement over DeepSeek R1 (F1 score of $72.1\%$) in political content detection. In bias/discrimination detection, \caro outperforms LLaMA-Guard-3-8B ($58.4\%$ F1 ) by $39.0\%$. The experimental results demonstrate that \caro's overall F1 score (89.2) represents an 11.5\% improvement over the second-best model (Full SFT, 80.0), validating the advancement of this method in content safety applications.
This study systematically evaluates the performance of various models in content moderation tasks (Table \ref{tab:overall}). Our proposed \caro method demonstrates significant advantages across all test categories, particularly excelling in tasks with ambiguous rules such as politically harmful content detection (F1 score of 89.5\%) and biased language identification (F1 score of 81.2\%). Compared to baseline models, \caro shows a +77.5 percentage point improvement over LLaMA-Guard-3-8B (12.0\%) and a +16.8 percentage point improvement over DeepSeek R1 (72.7\%) in political content detection. In biased detection, \caro outperforms LLaMA-Guard-3-8B (45.7\%) by +35.5 percentage points. The experimental results demonstrate that \caro's overall F1 score (89.2) represents a +5.0 percentage point improvement over the second-best model (Naive SFT, 84.2), confirming the effectiveness of this method for content moderation.

Additionally, we compared our approach against the recently proposed Class-RAG method \cite{chen2024class}, which leverages RAG to boost performance. While Class-RAG demonstrates improvements over standard base models, particularly on general cases, it still falls significantly short in the more difficult subcategories. A likely limitation is that conventional RAG methods are restricted to referencing a static pool of existing samples, lacking the capacity to dynamically generate the most contextually relevant analogical cases for unseen test instances (See detail in Table \ref{tab:case}).

Moreover, we also conducted experiments with an alternative workflow (Agentic ICL in Table \ref{tab:overall}), where an auxiliary model first generates reference cases which are most related to the target sentence, and then the main model performs moderation by drawing analogies to these references without any end-to-end optimization. Our results show that  without joint optimization, the model struggles to learn high-quality analogical behaviors and how to effectively utilize these examples, leading to less robust moderation outcomes.

\subsection{Ablation Study}
\caro comprises three key components as shown in Figure \ref{fig:framwor}:

 (1) \textbf{Bootstrapped COAT}: The model initially generates analogical reasoning chains in a self-supervised manner. (2) \textbf{Chain-of-Analogy Refinement and SFT}: Since the initially generated reasoning chains may not always align with the ground-truth labels, we introduce a reflection and refinement stage guided by supervised signals. This step fine-tunes the reasoning process to improve alignment with true labels. (3) \textbf{Analogical Reasoning Optimization with DPO}: Finally, we incorporate  DPO to further reinforce the analogical reasoning process.

In this part, we conduct ablation experiments to dissect the effectiveness of each optimization strategy.

As shown in Table \ref{tab:eval_results}, each of these key steps leads to a considerable improvement in the overall F1 score, which underscores their importance in the proposed framework. Our phased analysis reveals systematic gains at each stage. Specifically, Initial SFT using DeepSeek R1-generated reasoning chains on Qwen2.5-7B increases overall F1 score from 64.3\% to 85.5\%, with F1 score of political-harm  content increasing  from 58.6\% to 87.2\%, validating the effectiveness of chain-of-analogy fine-tuning for semantic understanding.

Subsequent DPO alignment led to a modest improvement in F1 score, \textbf{while significantly increasing the Chain-of-Analogy Ratio (CoA Ratio), which is defined as the proportion of test cases containing explicit analogical reasoning, as quantified in Table \ref{tab:eval_results}.} Relative to the limited improvement in the F1 score, this further stimulates the model's emergent ability for analogical reasoning, making the analogical reasoning chains more explicit and consistent.

The ablation confirms that each stage contributes to the final performance, with the gains accumulating across the pipeline.

\subsection{Discussion}

\nosection{\caro on other benchmarks}
To further demonstrate the generalizability of our approach, we conducted experiments not only on our primary dataset which contains a high proportion of ambiguous samples, but also on several widely-used public datasets. Since the primary dataset is in Chinese and the public benchmarks are in English, we evaluate generalization separately for each language. The results below report English-language experiments, while the Chinese results appear in Table \ref{tab:overall}. Specifically, we used the training split of the Aegis dataset (6,753 valid training samples after cleaning) to train a Qwen2.5-7B model with our proposed method. After training, we evaluated the model both in-distribution on the Aegis test set and out-of-distribution on two additional datasets: OpenAI-Moderation \cite{markov2023holistic} and Toxic-Chat \cite{lin2023toxicchat}.
As shown by the results in Table \ref{tab:safety_eval}, our approach boosts the average F1 score on the in-distribution Aegis test set from $78.7\%$ to $87.1\%$, a substantial improvement. Our method also achieves notable gains on the out-of-distribution datasets, despite never being trained on any data from these sources.
These results suggest that the analogical reasoning learned by \caro transfers across datasets and languages.

\nosection{Inference Cost Analysis}
Generating analogical chains during decoding introduces additional tokens compared to standard reasoning. In our experiments with Qwen2.5-7B-Instruct, \caro produces on average 332.9 extra tokens per example, while achieving an average F1 improvement of 24.9\%. This corresponds to roughly 13.4 extra tokens per 1-point F1 gain. Compared with conventional RAG methods, \caro avoids the cost of encoding and searching an external database at inference time, since all analogical behavior has been internalized into the model parameters during training.

\section{Related Work}

\nosection{Retrieval-Augmented and Agentic Approaches}
Our approach is closely related to two recent research lines. \textbf{Class-RAG} \cite{chen2024class} directly integrates RAG with the model to enhance content moderation by injecting retrieved examples into the model's context. However, it cannot generate novel, adaptive analogies tailored to unseen test samples, as the analogical reasoning is constrained by the retrieval database. \textbf{Agentic-RAG paradigms} such as Search R1 \cite{jin2025search} combine rule-based rewards and reinforcement learning to let the model decide when and what to retrieve. While this introduces retrieval autonomy, the retrieved cases are still drawn from a fixed database and share similar limitations. In our early experiments, we also explored integrating analogical reasoning with end-to-end RL with GRPO \cite{guo2025deepseek}, but without carefully designed constraints, unconstrained RL struggled to guide the model toward effective analogical reasoning behaviors. We leave more advanced RL-based analogical reasoning as future work.

\nosection{Content Moderation with Prompting LLMs}
A straightforward approach leverages powerful LLMs directly as zero-shot or few-shot moderators, relying solely on carefully engineered prompts without any further model updates \cite{kumar2024watch}. For instance, models such as GPT-4 or Claude can be instructed to detect inappropriate or unsafe content by providing detailed moderation policies within the prompt. Such methods are attractive for their simplicity, adaptability, and low resource requirements. However, recent studies \cite{kumar2024watch} report that pure prompting often struggles with nuanced or adversarial cases, especially when moderation guidelines are complex or ambiguous.

\nosection{Content moderation based on LLM Post-training }
To address the limitations of above approaches, a growing body of work explores post-training methods to align LLMs for content moderation. Notable examples include the LLaMA Guard family \cite{inan2023llama}, fine-tuned from LLaMA2 \cite{touvron2023llama}, LLaMA3 and LLaMA3.1 \cite{grattafiori2024llama}, respectively. LLaMA Guard pioneered the use of dedicated LLM-based guardrails for human-AI interactions. Similarly, WILDGUARD \cite{han2024wildguard} advances the field by introducing the first LLM-based moderator to explicitly assess both response harms and refusal behavior, improving adversarial robustness and outperforming prior tools in jailbreak detection. Other recent systems, such as Aegis \cite{ghosh2024aegis}, MD-Judge \cite{li2024salad}, and ShieldGemma \cite{zeng2024shieldgemma}, are trained on diverse safety datasets and policies to provide more reliable binary or categorical harm assessments.

\section{Conclusion}
In this work, we introduced \caro, an analogical reasoning optimization framework designed to enhance content moderation accuracy and robustness. Through extensive experiments on both in-distribution and out-of-distribution datasets, \caro demonstrated significant improvements over both general-purpose and retrieval-augmented baselines. Case studies further illustrate that \caro's generated analogical references are more semantically aligned with the input, allowing for nuanced and context-aware moderation decisions. Our results highlight the effectiveness of analogical reasoning in reducing hallucinations and improving generalization in challenging moderation scenarios.
\section*{Limitations}

While \caro shows promising performance, several limitations remain. First, although \caro reduces reasoning hallucination, there is potential for the model to generate misleading analogies in edge cases, which could affect moderation reliability. Additionally, the current framework has been evaluated primarily on text-based content moderation tasks; extending \caro to multi-modal or real-time moderation settings warrants further investigation.

\section*{Acknowledgments}
This work was supported by the National Natural Science Funds for Young Scholar under Grant 62503336.

\section*{Ethical Statement}
We adhere to the ACL Code of Ethics and have taken every measure to ensure that our research complies with the ethical guidelines set forth. Our work does not involve human subjects, nor does it raise any ethical concerns related to privacy or data usage. All datasets used in our experiments are publicly available, and we ensure that their release and use adhere to proper data-sharing policies.
We have carefully selected the datasets and evaluation metrics to ensure fairness and transparency in our findings. No potential conflicts of interest, sponsorship biases, or ethical violations have been identified in our study.
We commit to maintaining the highest standards of research integrity, and we are open to addressing any ethical concerns that may arise during the review process.

\bibliographystyle{unsrtnat}
\bibliography{custom}

\appendix

\section{Appendix}
\label{sec:appendix}

\subsection{Example Prompts}
\label{sec:prompts}

\subsubsection{Analogical Chain Generation Prompt}
\label{subsec:analogical_prompt}
The prompt used for generating analogical reasoning chains (as referenced in Section 2.1) is:

\begin{quote}
\small
\textbf{Task:} Assume you are a professional content moderator. Assess the harmfulness of the following content and classify it into exactly one of these categories: [Political Harmful, Pornography, Violence, Bias, Gambling, Harmless].

\textbf{Instructions:}
\begin{enumerate}[leftmargin=*,nosep]
    \item Provide a detailed "Analysis Process" explaining the reasoning for classification using chain-of-thought
    \item Extract the "Harmful Content" from the sentence
    \item Provide the "Classification Result"
\end{enumerate}

\textbf{Format Requirements:}
\begin{itemize}[leftmargin=*,nosep]
    \item Response must follow this three-part format: 'Analysis Process', 'Harmful Content', 'Classification Result'
    \item You may reference "Example Cases" if available (note they may not always be accurate)
    \item Any referenced examples should be reflected in your final response
\end{itemize}
\end{quote}

\subsubsection{Reflection \& Refinement Prompt}
\label{subsec:reflection_prompt}
The prompt used for the reflection process (as referenced in Section 2.1) is:

\begin{quote}
\small
\textbf{Task:} Assume you are a professional content moderator. Assess the harmfulness of the following content and classify it into exactly one of these categories: [Political Harmful, Pornography, Violence, Bias, Gambling, Harmless].

\textbf{Instructions:}
\begin{enumerate}[leftmargin=*,nosep]
    \item Provide a detailed "Analysis Process" explaining the reasoning for classification using chain-of-thought
    \item Extract the "Harmful Content" from the sentence
    \item Provide the "Classification Result"
\end{enumerate}

\textbf{Additional Context:}
\begin{itemize}[leftmargin=*,nosep]
    \item Your original response was: \{item['response']\}
    \item Please reflect and provide a final answer
    \item Do not explicitly mention reflection in your final output
\end{itemize}

\textbf{Format Requirements:}
\begin{itemize}[leftmargin=*,nosep]
    \item Response must follow three-part format: 'Analysis Process', 'Harmful Content', 'Classification Result'
    \item You may reference "Example Cases" if available (note they may not always be accurate)
    \item Any referenced examples should be reflected in your final response
\end{itemize}
\end{quote}

\subsection{Dataset Details}
\label{sec:dataset}

\subsubsection{Category Definitions}
\label{subsec:category_defs}
The dataset contains 6 categories of harmful content (as referenced in Section 3.2):

\begin{description}[leftmargin=2em,style=nextline]
\item[Politics] Content that violates political compliance requirements or contains sensitive ideological elements across different cultural contexts.
\item[Pornography] Sexually explicit content intended to cause arousal.
\item[Violence] Content that promotes, glorifies, or threatens physical harm.
\item[Bias] Content containing discriminatory generalizations or stereotypes about specific groups.
\item[Gambling] Content promoting or facilitating gambling activities.
\item[Harmless] Content that does not fall into any harmful categories.
\end{description}

\subsubsection{Dataset Statistics}
\label{subsec:dataset_stats}
We repartition the original dataset to balance category distributions, mitigating training bias and ensuring fairer evaluation, as shown in table \ref{tab:dataset_splits}
\begin{table}[t]
\centering
\caption{Dataset Splits and Distribution}
\label{tab:dataset_splits}
\begin{tabular}{lrrr}
\toprule
\textbf{Category} & \textbf{Total} & \textbf{Train} & \textbf{Test} \\
\midrule
Politics & 4,378 & 1,200 & 250 \\
Pornography & 2,519 & 1,200 & 250 \\
Violence & 1,472 & 1,200 & 250 \\
Bias & 1,978 & 1,200 & 250 \\
Gambling & 2,079 & 1,200 & 250 \\
Harmless & 2,129 & 1,200 & 250 \\
\midrule
\end{tabular}
\end{table}

\subsubsection{Data Preprocessing}
The dataset was split into 7,200 training samples and 1,800 test samples at an 8:2 ratio, ensuring balanced distribution across categories. The raw data underwent multiple processing steps to enhance quality and diversity: Since duplicated samples cause models to overfit spurious patterns without improving generalization, we employed pretrained text encoders to generate sentence embeddings, performed cosine-similarity-based clustering for each category and selected the most representative sample from each cluster. This rigorous deduplication prevents the model from being misled by repetitive but uninformative data variations.
\begin{table*}[htbp]
    \centering
    \caption{Case Study: \caro vs. baseline reference retrieval in content moderation.}
    \label{tab:case}
    \scriptsize
    \begin{tabular}{p{5cm}p{0.7cm}p{3cm}p{4cm}}
        \toprule
        \textbf{Sentence} & \textbf{Label} & \textbf{Reference (\caro generated)} & \textbf{Reference(Qwen2.5-7B-Instruct)} \\
        \midrule
        I feel that whenever Africa is mentioned, it is instinctively linked to \textcolor{red}{poverty and disease}, and Dubai is associated with \textcolor{red}{wealth}. It's similar to how foreigners immediately think of \textcolor{red}{Chinese Kung Fu and giant pandas} upon hearing about China. & Bias & "\textcolor{red}{Black people} are all \textcolor{red}{lazy}", "\textcolor{red}{Indians} are all \textcolor{red}{liars}", "\textcolor{red}{Regional discrimination} against \textcolor{red}{Northeast people}" & "At that time, \textcolor{gray}{Asians inspired Africa and Latin America}. In Egypt, yellow - skinned people were \textcolor{gray}{respected} for defeating white - skinned people." \\
        \midrule
        Just talk about the roommate. Why bring up the roommate from Inner Mongolia? What's the point? This can easily come across as overgeneralizing and unfriendly... & Harmless & "Why don't you say it's from Guizhou?", "Isn't it from Jiangxi?" & "There are many \textcolor{gray}{black intermediaries} and \textcolor{gray}{annoying second-landlords from Northeast China} in Beijing", "Aren't \textcolor{gray}{people from Northeast China} generally straightforward? Why is it like this...?" \\
        \midrule
        The accent of the Chinese Korean ethnic group is closer to the Korean language of North Korea rather than that of South Korea, right [Surprise] & Harmless & "The Chinese pronunciation of North Koreans is related to the Korean language" & "Don't be \textcolor{gray}{overly feminist} about everything", " \textcolor{gray}{South Korean international students} have extremely \textcolor{gray}{low quality}", "Some \textcolor{gray}{South Koreans} indeed have extremely \textcolor{gray}{low quality"} \\
        \midrule
        The Anti Bar leans to the left. Why are there so many posts in the Anti Bar that \textcolor{red}{look down on black people}? Isn't \textcolor{red}{discriminating against black people} a typical characteristic of the right wing? & Bias & "Why are \textcolor{red}{black people discriminated} against?", "Black people are \textcolor{red}{all lazy}" & "At that time, \textcolor{gray}{Asians} inspired \textcolor{gray}{Africa and Latin America}. In Egypt, \textcolor{gray}{yellow-skinned} people were respected for defeating \textcolor{gray}{white-skinned} people." \\
        \bottomrule
    \end{tabular}

\end{table*}
\subsection{Why \caro works: a view from case study}
Table \ref{tab:case} presents several examples where our method successfully classified cases that the ClassRAG \cite{chen2024class} approach failed to handle. By examining these specific cases, we can better understand why our \caro framework is effective.
Taking the first row as an example, \caro generates reference cases that directly highlight biased generalizations and stereotyping (key information is highlighted with \textcolor{red}{red} color), which are semantically aligned with the original sentence's pattern of associating entire groups with certain traits. This analogy enables the model to correctly flag the sentence as bias. In contrast, references retrieved in static training set is only tangentially related and fails to capture the core issue of stereotyping, leading to misclassification . In the second row, \caro provides references that mirror the structure of the original statement, questioning why a specific region is mentioned but without introducing negative stereotypes. This helps the model recognize that the statement, while potentially awkward, is not inherently biased. The RAG method, however, retrieves references with explicit negativity or discrimination, potentially leading the model to over-moderate harmless content.

Across these cases, \caro consistently surfaces analogies that are structurally and semantically aligned with the key issues in the input, whether it is discrimination, stereotyping, or neutrality. This enables robust, context-sensitive moderation. In contrast, baseline methods often surface off-topic or misleading references, resulting in suboptimal decisions.

\end{document}